\documentclass[sn-mathphys]{sn-jnl}% Basic Springer Nature Reference Style/Chemistry Reference Style

\jyear{2022}%
\usepackage{subcaption}
\theoremstyle{thmstyleone}%
\theoremstyle{thmstyletwo}%

\theoremstyle{thmstylethree}%
\newcommand{\ours}{TC-CaRTS}
\begin{document}

\title[\ours{}]{Rethinking Causality-driven Robot Tool Segmentation with Temporal Constraints}


\author*[1]{\fnm{Hao} \sur{Ding}}\email{hding15@jhu.edu}

\author[2]{\fnm{Jie Ying} \sur{Wu}}

\author[1]{\fnm{Zhaoshuo} \sur{Li}}

\author*[1]{\fnm{Mathias} \sur{Unberath}}\email{unberath@jhu.edu}

\affil*[1]{\orgdiv{Department of Computer Science}, \orgname{Johns Hopkins University}, \orgaddress{\street{3400 N. Charles St}, \city{Baltimore}, \postcode{21218}, \state{MD}, \country{USA}}}

\affil[2]{\orgdiv{Department of Computer Science}, \orgname{Vanderbilt University}, \orgaddress{\street{2201 West End Ave}, \city{Nashville}, \postcode{37235}, \state{TN}, \country{USA}}} 

% \affil[3]{\orgdiv{Department}, \orgname{Organization}, \orgaddress{\street{Street}, \city{City}, \postcode{610101}, \state{State}, \country{Country}}}


\abstract{
\textbf{Purpose:} Vision-based robot tool segmentation plays a fundamental role in surgical robots and downstream tasks. 
%In surgical scenarios, performance and robustness are two vital aspects of success and safe surgery. Recent deep learning architectures solely focus on reporting the performance on benchmark datasets but lack discussion of the challenges in robustness against domain shift. 
CaRTS, based on a complementary causal model, has shown promising performance in unseen counterfactual surgical environments in the presence of smoke, blood, etc. However, CaRTS requires over 30 iterations of optimization to converge for a single image due to limited observability.

\textbf{Method:} To address the above limitations, we take temporal relation into consideration and propose a temporal causal model for robot tool segmentation on video sequences. We design an architecture named Temporally Constrained CaRTS (\ours{}). \ours{} has three novel modules to complement CaRTS -- temporal optimization pipeline, kinematics correction network, and spatial-temporal regularization. 

\textbf{Results:} 
Experiment results show that \ours{} requires much fewer iterations to achieve the same or better performance as CaRTS. \ours{} also has the same or better performance in different domains compared to CaRTS. All three modules are proven to be effective.

\textbf{Conclusion:} 
We propose \ours{}, which takes advantage of temporal constraints as additional observability. We show that \ours{} outperforms prior work in the robot tool segmentation task with improved convergence speed on test datasets from different domains. 
}




\maketitle

\section{Introduction}\label{sec1}

With the widespread application of surgical robots and the growing demand for autonomous surgery, vision-based robot tool segmentation plays a fundamental role in robot perception~\cite{luis2017ToolNet,jin2019ITP,shevets2018AIS,pakhomov2019DRLISRS,mobarakol2019rtis,qin2019SICF,zhao2021one-to-many,su2018real,da2019self,emanuele2020synthetic}. In surgical scenes, performances and robustness of segmentation algorithms are both important aspects for the success and safety of downstream operations. Various feed-forward networks and machine-learning techniques designed for semantic/instance segmentation have achieved promising performance~\citep{olaf2015unet, chen2018deeplabv3+, he2017maskrcnn, chen2019htc, ding2021dsc, liu2021Swin, WangSCJDZLMTWLX19,cheng2021stcn}. However, their performance does not generalize when tested on data from different domains~\cite{nathan2021robustness}. To improve robustness, some effort has already been made ~\cite{motrovic2021RLIC, ouyang2021causalDomain, zhang2020cvrnn, liu2021causalSemantic,ding2022carts}. CaRTS~\cite{ding2022carts}, designed from a complementary causal model, shows a promising and robust performance when tested on counterfactual surgical environments for robot tool segmentation. However, optimization from an image-wise perspective faces limited observability. This limitation makes CaRTS hard to optimize.

\begin{figure}[t]%
\centering
\includegraphics[width=\textwidth]{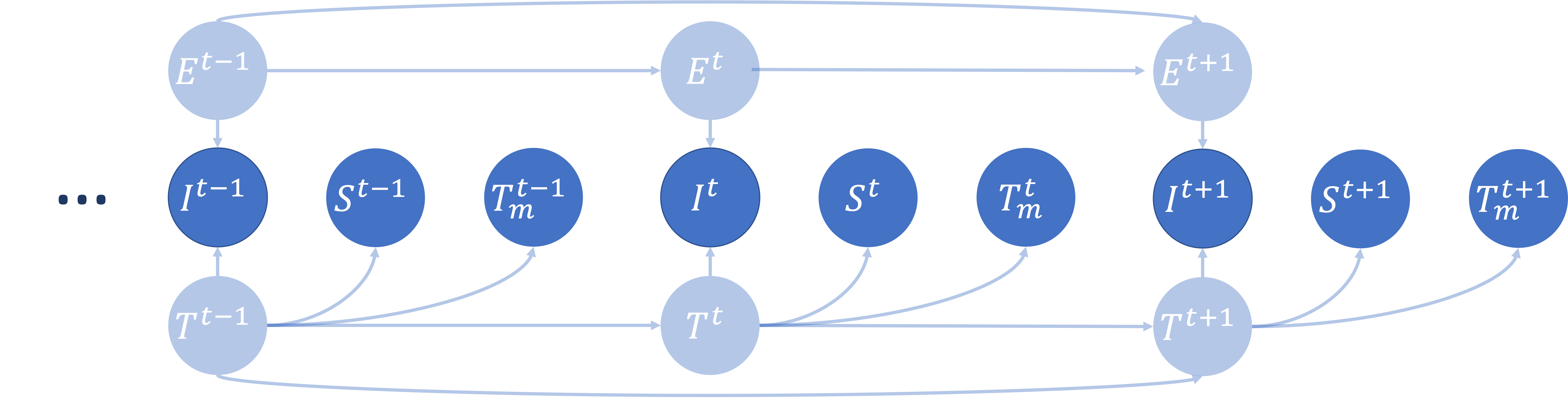}
\caption{Illustration of the temporal causal model for robot tool segmentation task. Arrow lines mean direct causal effect. 
%Dashed lines mean direct causal effects that can be ignored under certain assumptions. Light blue represents unobserved factors; dark blue nodes represent observed factors. 
At timestamp $t$, We note $\mathbf{I^t}$ for image, $\mathbf{E^t}$ for environment, $\mathbf{S^t}$ for segmentation, $\mathbf{T^t}$ for true robot kinematics and camera poses, $\mathbf{T^t_m}$ for measured robot kinematics and camera poses}
\label{causal_model}
\end{figure}

In order to alleviate this issue, we propose a temporal causal model which frames robot tool segmentation along a sequence. This temporal causal model is shown in Fig.~\ref{causal_model}. We use the same idea from CaRTS where images $\mathbf{I^t}$ and segmentation $\mathbf{S^t}$ at timestamp $t$ are directly determined by all unobserved robot and camera parameters $\mathbf{T^t}$, and the environment $\mathbf{E^t}$ at the same timestamp $t$. We assume occlusion has no effect on segmentation and there is no interaction between tools and environments. As the model describes, $\mathbf{T^{t+1}}$ at timestamp $t+1$ are directly determined by $\mathbf{T^{t}}$, $\mathbf{T^{t-1}}$, and all $\mathbf{T}$s in the past. Exploring this temporal causal effect might provide temporal constraints that are effective to deal with the issues mentioned above. 

We explore the underlining temporal constraints in this temporal causal model. The first direction is to differentiate time-variant and time-invariant factors in $\mathbf{T}$ and optimize them differently according to their property. Kinematics, with notation $\mathbf{K}$, is representative of time-variant factors. Modeling the causal effect of $\mathbf{K}$s by $P(\mathbf{K^t} \mid \mathbf{K^{t-1}}, \mathbf{K^{t-2}}, \dots)$ can be a promising direction since the model can learn motion property from the previous trajectory to provide constraints. Another direction for the time-variant factors is making the spatial-temporal smoothness assumption, which assumes the measurement error and the inter-frame motion is small. This assumption should be safe to make since the difference between two adjacent timestamps is small and the speed of the robot is limited. Inspired by these directions, we propose temporally constrained CaRTS (\ours{}) architecture. \ours{} has three novel modules upon CaRTS. The first is a temporal optimization pipeline that enables the optimization of time-invariant factors like base configuration. The second is a kinematics correction network (KCN) that models the temporal causal effect for kinematics. The third is a spatial-temporal regularization based on the spatial-temporal smoothness assumption.

Our experiments show that \ours{} effectively reduced the required iterations to achieve the same or better segmentation performance compared to CaRTS. Ablation studies indicate this improvement comes from the temporal constraints utilized by all three proposed modules. The code will be released.

In summary, the main contributions of this paper are as follows: (1) Modeling temporal causal relations and exploring potential direction temporal constraints in this model. (2) Proposing \ours{} that utilize the temporal constraints in (1) that requires fewer iterations to achieve the same or better performance compared to CaRTS.

% Taking advantage of the temporal information, our new pipeline achieves ?? times speed compared to the primitive CaRTS pipeline and ?? initial dice score before the optimization of each frame while retaining the robustness property of CaRTS.
\section{Related Work}\label{sec2}

\paragraph{Robot Tool Segmentation:} Both image-wise and video-wise semantic or instance segmentation have already been a maturely-developed area. Feed-forward networks, e.g. ~\citep{olaf2015unet, chen2018deeplabv3+, he2017maskrcnn, chen2019htc, ding2021dsc, liu2021Swin, WangSCJDZLMTWLX19,cheng2021stcn}, are on a dominating stand. Their variants ~\cite{luis2017ToolNet,pakhomov2019DRLISRS,shevets2018AIS,jin2019ITP,mobarakol2019rtis,zhao2021one-to-many} are also the state-of-the-art on robot tool segmentation. At the same time, increasing efforts have been made to incorporate other available information, e.g. geometric information~\cite{AllanOHKS18, LiLDDCTU21sttr, YeZGY16rt3dtracking} and kinematics~\cite{su2018real, da2019self}, with visual input to improve performance~\cite{qin2019SICF,su2018real,da2019self} or robustness~\cite{emanuele2020synthetic,ding2022carts} for robot tool segmentation.

\paragraph{Causality in Computer Vision:} Causality has been receiving increasing attention in computer vision research, especially medical area. Some researchers use ideas from causal inference to design feature representation learning methods~\cite{motrovic2021RLIC, ouyang2021causalDomain, zhang2020cvrnn, liu2021causalSemantic} for domain generalization. Some researchers use the concept of counterfactual for generative models~\cite{jacob2021causalMRI, nick2020DSCM}. Lenis et al.~\cite{dimitrios2020Domainaware} use this concept for the interpretability of medical image classifiers. Some researchers focus on posing the underlying causal model of the vision task~\cite{castro2020causality, ding2022carts}.

\paragraph{CaRTS:} Ding et al.~\cite{ding2022carts} have proposed a novel causal model where the segmentation is directly determined by the robot kinematics, camera poses, and the environment instead of the observed image. Based on this causal model they design CaRTS architecture that iteratively optimizes feature similarity between rendered images and observed images w.r.t the measured kinematics to estimate true kinematics. The final segmentation is the rendered silhouette of the robot model given the estimated kinematics. CaRTS achieves outstanding robustness across testing domains compared to other feed-forward networks. However, limited observability in image-wise optimization makes CaRTS hard to achieve real-time inference. Our temporal causal model and \ours{} architecture incorporate temporal constraints that are intuitively helpful for this limitation. 

\section{Method}\label{sec3}

We propose the \ours{} architecture based on our temporal causal model. The basic modeling of \ours{} is similar to CaRTS but has three novel modules. We introduce the temporal optimization pipeline first and then introduce KCN and spatial-temporal regularization.

\begin{figure}[t]%
\centering
\includegraphics[width=\textwidth]{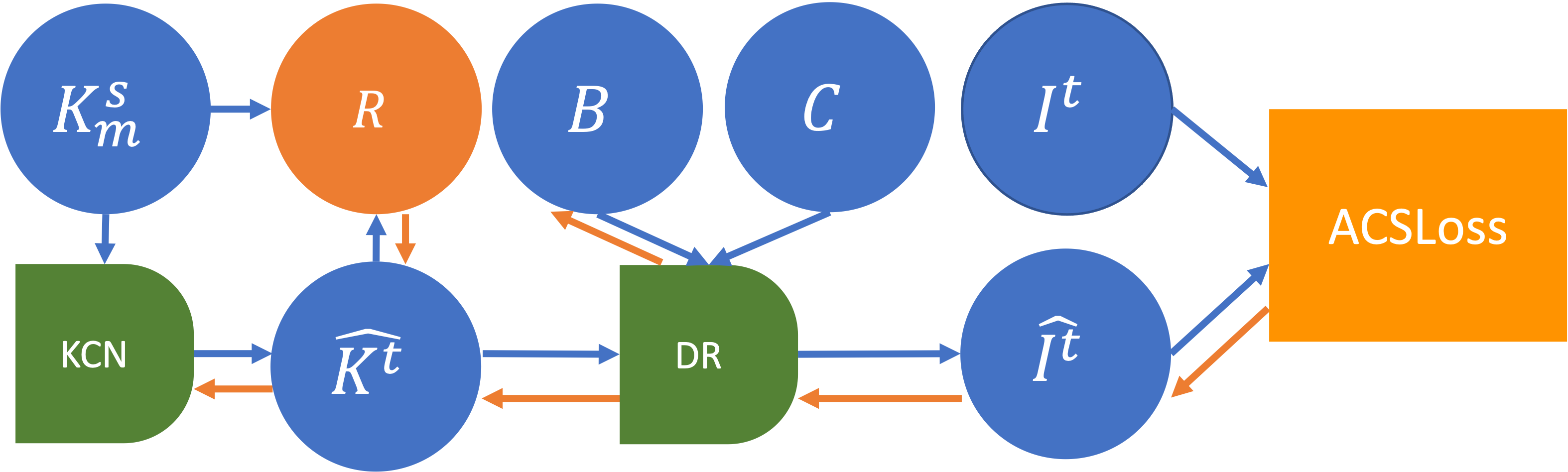}
\caption{Illustration of the overall architecture of \ours{}. $\mathbf{K^s_m}$ denotes the robot kinematics of a time segment $\mathbf{s} = \{t, t-1, \dots, t-n-1\}$. $\mathbf{\hat{K^t}}$ denotes corrected kinematics output by KCN. $\mathbf{R}$ denotes spatial-temporal regularization term. $\mathbf{B}$ denotes robot base configuration.  $\mathbf{C}$ denotes all other parameters that are required for rendering, e.g. camera configurations, mesh models, DH parameters, etc. $\mathbf{I^t}$ and $\mathbf{\hat{I^t}}$ denotes observed and rendered image respectively. Solid blue lines in this figure represent information flow and solid orange lines represent gradient flow. DR stands for differentiable rendering.}
\label{ti_carts}
\end{figure}

\subsection{Temporal Optimization Pipeline}

The temporal optimization pipeline is illustrated in Fig.\ref{ti_carts}. Notations are described in the caption of Fig.\ref{ti_carts}. The optimization objective function in \ours{} is the spatial-temporal regularized ACSLoss between deep feature maps extracted from the observed image $\mathbf{I^t}$ and rendered image $\mathbf{\hat{I^t}}$. ACSLoss calculates an attentional cosine similarity between feature maps. CaRTS use pre-trained U-Net to extract feature maps from observed image and a hybrid image made up of rendered robot tool and the average background from the training dataset. Optimizing ACSLoss between these two feature maps aligns the calculated robot configuration to the observed robot. Since the robot tool is differentially rendered, the gradient can be backpropagated to the input kinematics. Thus, gradient descent can be performed to correct the measurement error. More details about the optimization pipeline can be found in \cite{ding2022carts}

Different from the image-wise CaRTS pipeline, the temporal optimization pipeline differentiates the time-variant factor and time-invariant factor. In \ours{}, time-variant means robot kinematics sequence $\mathbf{K_m^s}$ and time-invariant factor means robot base configuration $\mathbf{B}$. Other factors $\mathbf{C}$ remain constant during optimization. The optimization objective function can be written as Eq.~\ref{eq_optimize_object} where $\theta$ is KCN's weights.

\begin{equation}
    \arg\min_{\theta, \mathbf{B}} ACSLoss(\mathbf{\widehat{K^t}}, \mathbf{B}, \mathbf{C}) + \mathbf{R}
\label{eq_optimize_object}
\end{equation}

During optimization, we alternatively perform gradient descent for $\theta$ and $\mathbf{B}$. At each timestamp $t$, we first calculate ACSLoss, backpropagate gradient, and perform gradient descent for $\theta$ in KCN for $k$ iterations to learn the temporal relation of kinematics for KCN. Then, we freeze KCN and repeat the above optimization process for $\mathbf{B}$ for one iteration. The final corrected kinematics $\mathbf{\hat{K^t}}$, base configuration $\mathbf{\hat{B}}$, and $\mathbf{C}$ are used to render the predicted segmentation $\mathbf{\hat{S^t}}$. 

\subsection{Kinematics Correction Network}

The Kinematics Correction Network (KCN) is an MLP network $F_{\theta}$. The input kinematics $\mathbf{K_m^s}$ is a $n \times d$ matrix, $d$ is the dimension of the kinematics, and $n$ is the number of timestamps. $PE(n)$ is the positional encoding for each timestamp. The output is the corrected kinematics $\mathbf{\hat{K^t}}$ that estimates the true kinematics. KCN can be expressed as Eq.~\ref{equ_k_t}:

\begin{equation}
    \mathbf{\widehat{K^t}} = \mathbf{K_m^s} + F_{\theta}(\mathbf{K_m^s} + PE(n))
\label{equ_k_t}
\end{equation}

\subsection{Spatial-temporal Regularization}
The spatial-temporal regularization is based on the spatial-temporal smoothness assumptions that (a) the measurement error is small and (b) the inter-frame motion between consecutive timestamps is small. As we use joint angles as kinematics to optimize, we use the L2 norm to regularize them. The Regularization term can be written as:

\begin{equation}
    R = \lambda_1\frac{1}{d}\sum_{i=1}^d (\mathbf{\widehat{K^{t,i}}} - \mathbf{K^{t,i}}) ^ 2 + \lambda_2\frac{1}{d}\sum_{i=1}^d (\mathbf{\widehat{K^{t,i}}} - \mathbf{\widehat{K^{t-1,i}}}) ^ 2
\label{equ_r}
\end{equation}

\noindent where $i$ represents the dimension indices of the kinematics $\mathbf{\widehat{K^{t, i}}}$ calculated via Eq.~\ref{equ_k_t}. $R$ denotes the whole regularization term, $\lambda_1,\lambda_2$ are hyperparameters for adjusting regularization strength of each term. $\mathbf{(\widehat{K^{t, i}}} - \mathbf{K^{t, i}} )^ 2$ in the first term regularizes according to assumption (a) and $ (\mathbf{\widehat{K^{t, i}}} - \mathbf{\widehat{K^{t-1, i}}}) ^ 2$ in the second term regularizes according to assumption (b).

\section{Experiment}\label{sec4}

\begin{figure}[t]%
\centering
\includegraphics[width=\textwidth]{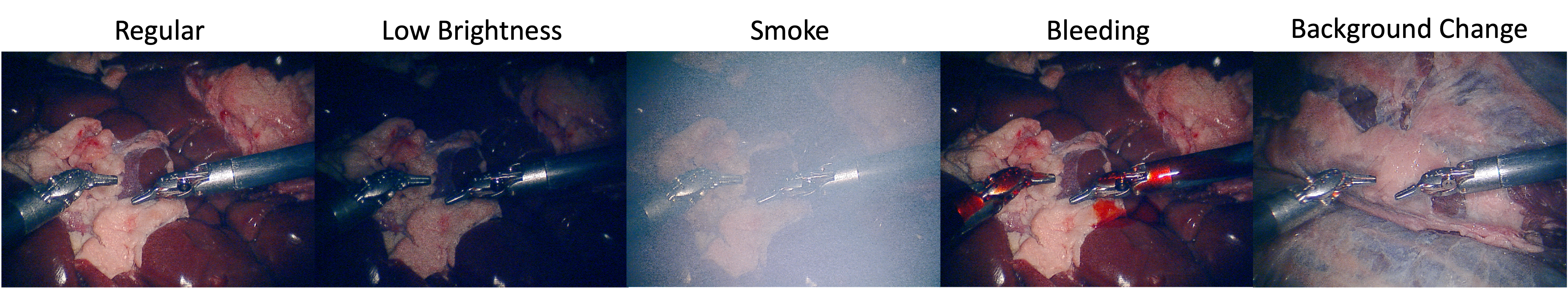}
\caption{Example of the counterfactual images from different domains at the same timestamp}
\label{dataset}
\end{figure}

We perform the robot tool segmentation experiment on the dataset from CaRTS\cite{ding2022carts}. The dataset has nine videos (seven for training, one for validation, and one for testing). Each video contains 400 frames. Each frame has its corresponding kinematics for both patient-side manipulators (PSMs). All videos were recorded under the same camera setting and robot base configuration which are roughly measured before recording. Robot motion recorded in all videos is in free space and no occlusion exists. We train the U-Net~\cite{olaf2015unet} feature extractors and all baseline models on the training dataset which only contains videos recorded from one domain without any corruption. We call this domain the regular domain. The validation and test dataset contain counterfactual videos that are recorded on other domains, e.g. smoke, bleeding, etc. Examples of the counterfactual image from different domains are shown in Fig.~\ref{dataset}.

In our experiment, we use Dice as the metric for measuring segmentation performance. We first explore the inference speed improvement of our \ours{} architecture compared to CaRTS on the test dataset. We also compare the performance of \ours{} to other deep learning algorithms on the test dataset. Then we present ablation studies on the validation dataset.

\subsection{Implementation Details}

In our experiment, we use the same implementation setting from CaRTS. For KCN, we choose input length $n = 5$ and optimize all six joint angles and one tool angle of the two PSMs which makes the dimension of the kinematics $d = 14$, KCN has five hidden layers with $32, 64, 128, 128, 64, 32$ channels. For spatial-temporal regularization we set $\lambda_1 = 10$ and $\lambda_2 = 1$. We use Adam optimizer with a learning rate of $5 \times 10^{-5} $ / $3 \times 10^{-6} $ for $\theta$ / $\mathbf{B}$. All baselines are trained for $50$ epochs with smoke augmentation. All of the experiments run on a single NVIDIA GeForce RTX 3090 graphic card. 

\begin{figure}[t]
\begin{subfigure}[b]{0.49\textwidth}
         \centering
         \includegraphics[width=\textwidth]{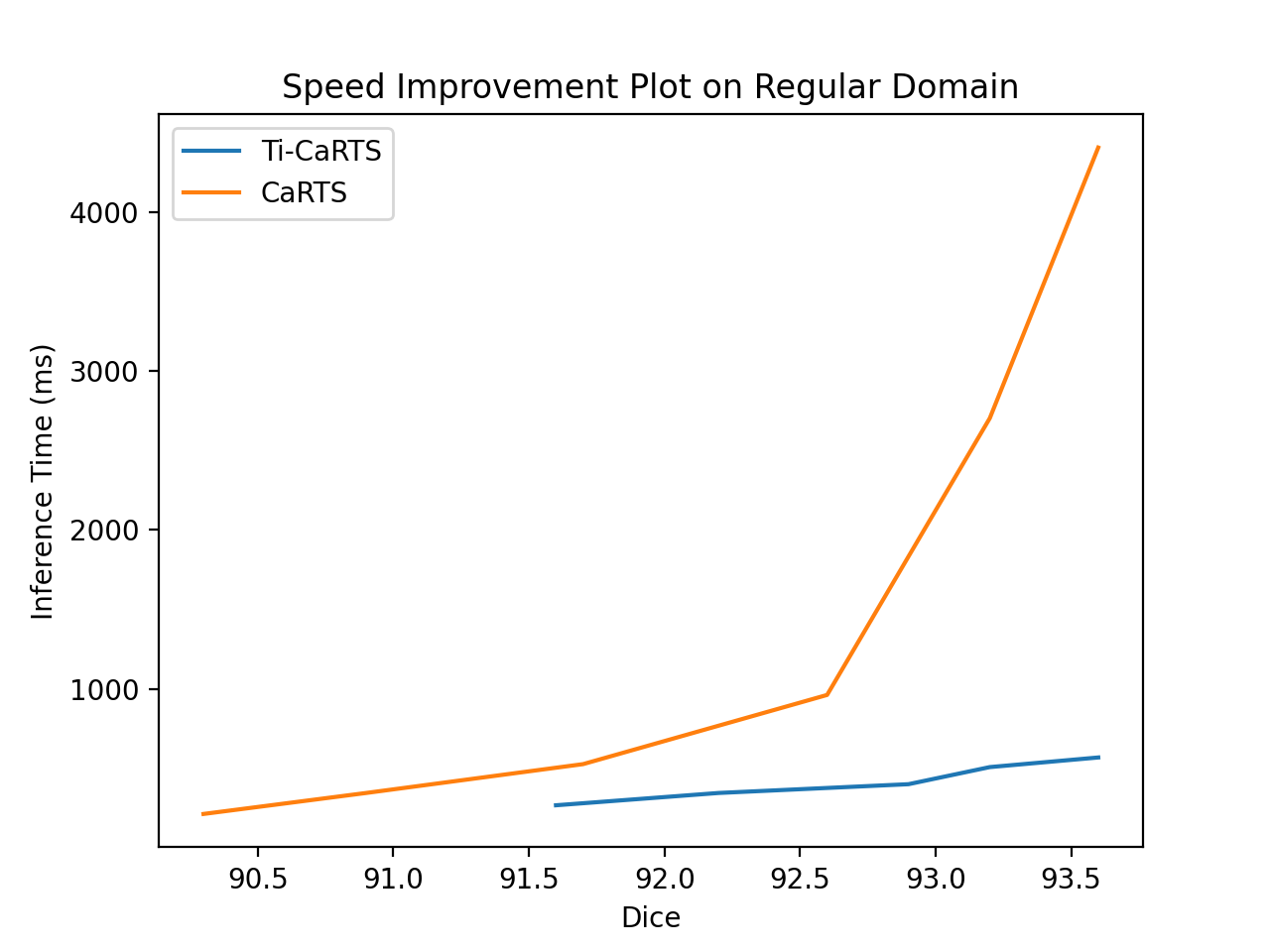}
         \caption{Plot on regular domain}
         \label{speed_improvement_regular}
     \end{subfigure}
 \hfill
 \begin{subfigure}[b]{0.49\textwidth}
         \centering
         \includegraphics[width=\textwidth]{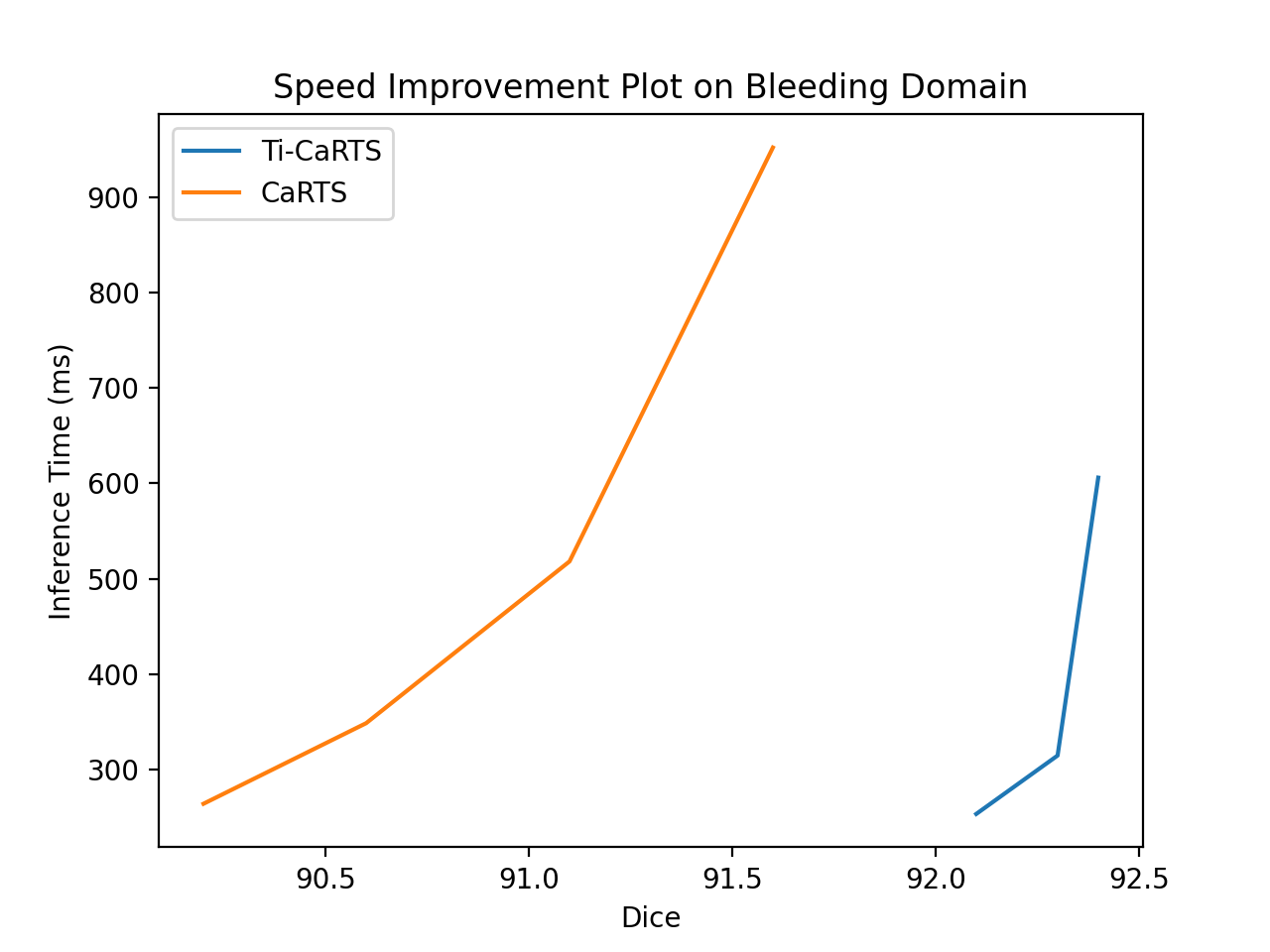}
         \caption{Plot on bleeding domain}
         \label{speed_improvement_bleeding}
     \end{subfigure}
\caption{Results of the speed improvement experiment}
\label{speed_improvement}

\end{figure}

\subsection{Speed Improvement} 

To measure speed improvement, we draw inference time vs performance plots for both CaRTS and \ours{} when inferring with different iteration times $k$ per frame. We choose $k = 1, 2, 3, 5, 10, 30, 50$ and test on the regular domain and bleeding domain. The quantitative results are in Tab.~\ref{speed_improvement_table} and the corresponding plot is shown in Fig~\ref{speed_improvement}. From the results, both CaRTS and \ours{} converge at a dice score of $93.6$ on the regular domain while \ours{} requires much fewer iterations ($5$ vs $50$) than CaRTS. On the bleeding domain, \ours{} after a single iteration outperformed CaRTS optimized with $10$ iterations. Although \ours{} is not real-time yet, it decreases the number of iterations for the optimization to converge. This is a significant step towards real-time inference.

\begin{table}[t]
\centering
\caption{Quantitative Results of the Speed Improvement}
\resizebox{\columnwidth}{!}{%
\begin{tabular}{|c|c|c|c|c|c|c|}
\hline
 Regular &  $k =$ 1 & 3  & 5 & 10 & 30 & 50\\
\hline
 CaRTS & $90.3 \pm 3.8$ & $90.9 \pm 3.8$ & $91.7 \pm 3.6$ & $ 92.6 \pm 3.2$ &  $93.2 \pm 3.6$ & $93.6 \pm 2.9$ \\
  inference time & 213ms & 345ms & 526ms & 961ms &  2702ms & 4405ms \\
\ours{} & $91.6 \pm 3.9$ & $93.2 \pm 2.8$ & $93.6 \pm 2.7$ & $ 93.6 \pm 2.7$ &  - & - \\
inference time & 267ms & 415ms & 568ms & - &  - & - \\
\hline
\hline
 Bleeding & $k =$ 1 & 3  & 5 & 10 & 30 & 50\\
\hline
 CaRTS & $90.2 \pm 3.8$ & $90.6 \pm 3.7$ & $91.1 \pm 3.7$ & $ 91.6 \pm 4$ &  $91.6 \pm 4$ & - \\
 inference time & 263ms & 348ms & 518ms & - &  - & - \\
 \ours{} & $92.1 \pm 3.1$ & $92.3 \pm 3$ & $92.4 \pm 2.9$ & - &  - & - \\
 inference time & 253ms & 314ms & 606ms & - &  - & - \\
\hline
\end{tabular}
}
\label{speed_improvement_table}
\end{table}

\subsection{Performance on Robot Tool Segmentation} 

In this experiment, we compare the tool's Dice score of \ours{} to CaRTS~\cite{ding2022carts}, image-based baselines including HRNet~\cite{WangSCJDZLMTWLX19}, Swin Transformer~\cite{liu2021Swin}, and method by Colleoni et al~\cite{emanuele2020synthetic} and a video-based baseline STCN~\cite{cheng2021stcn}. All baselines are trained with simulated smoke augmentation. As Tab.~\ref{RTS_Results} shows, all the feed-forward network-based methods perform well on the regular domain and they can achieve real-time inference. However, their performances deteriorate significantly in other domains that are unseen in the training dataset. \ours{} retains comparable performance as CaRTS on all domains. 

\begin{table}[t]
\centering
\caption{Robot Tool Segmentation Results}
\resizebox{\columnwidth}{!}{%
\begin{tabular}{|c|c|c|c|c|c|c|}
\hline
 & Regular  & Low Brightness & Bleeding & Smoke & BG Change & FPS \\
\hline
Colleoni's & $94.9 \pm 2.7$ &$87.0 \pm 4.5$ & $55.0 \pm 5.7$& $59.7 \pm 24.7$& $75.1 \pm 3.6$ &  35.8\\
HRNet& $ 95.2 \pm 2.7$ & $ 86.3 \pm 3.9 $ & $ 56.3 \pm 16.4$ & $ 77.2 \pm 23.6$ & $92.1  \pm 4.6 $ & 15.6\\
Swin Transformer&  $95.0\pm 5.5$  &  $ 93.0\pm 5.5$ &  $76.5 \pm 9.0$    &  $82.4\pm 17.0$  & $94.8\pm 5.3$ & 24.4\\
STCN & $92.2\pm 2.7$  & $64.3 \pm 6.9$ & $30.8\pm 10.4$& $69.2 \pm 26.5$ & $84.0\pm 5.6$ & 27.5\\
% Algorithm 2 &   & & & & & & \\
CaRTS&  $93.4 \pm 3.0$ & $92.4 \pm 3.1$ & $90.8 \pm 4.4$& $91.6 \pm 4.7$ & $92.3 \pm 4.8$ & 0.37\\
\ours{}&  $93.6 \pm 2.7$ & $92.3 \pm 3.3$& $92.2 \pm 3.3$ & $91.9 \pm 4.5$ &  $92.5 \pm 3.1$ & 1.76\\
\hline
\end{tabular}%
}
\label{RTS_Results}
\end{table}

\subsection{Ablation Study}

We perform ablation studies on the regular domain of the validation dataset to explore the effectiveness of all modules and design choices. 

\paragraph{Effectiveness of all modules:} We explore the effectiveness of the modules by adding them to the CaRTS architecture. From Tab.~\ref{ablation_components}, we find that using the temporal optimization pipeline can improve performance. Adding KCN can also improve performance when $k$ is small But when $k$ becomes larger, it might overfit some frames and fail to generalize without spatial-temporal regularization which is shown as the result when $k = 10$. The spatial-temporal regularization not only improves the performance when $k$ is small and accelerates the convergence but also stabilizes the optimization process of KCN for larger $k$.

\begin{table}[t]
\centering
\caption{Results of the ablation study for the effectiveness of all modules}
\resizebox{\columnwidth}{!}{%
\begin{tabular}{|c|c|c|c|c|c|c|c|c|}
\hline
CaRTS & Ti Optim   & KCN  & ST Reg & $k=1$ & $k=3$ & $k=5$ & $k=10$ \\
\hline
\checkmark & & & & $ 89.4 \pm 2.8$ & $ 90.8 \pm 3.0$ & $91.2 \pm 3.0$ &  $91.9 \pm 3.7$\\
\checkmark & \checkmark &  &  & $ 90.4 \pm 2.8$ & $ 91.5 \pm 2.8$  &  $91.9 \pm 3.0$ & $ 92.1 \pm 2.4$ \\
\checkmark & & \checkmark & &  $90.6 \pm 3.2$ & $91.1 \pm 3.0$& $91.5 \pm 2.7$ & $54.3 \pm 43.0$ \\
\checkmark & & \checkmark & \checkmark &  $ 91.1 \pm 2.9$ & $ 91.9 \pm 2.6$  &  $92.2 \pm 2.9$ & $ 92.4 \pm 2.7$\\

\checkmark & \checkmark & \checkmark & \checkmark & $91.4 \pm 3.1$ & $92.5 \pm 2.6$ & $92.7 \pm 2.6$ & $ 92.8 \pm 2.4$\\
\hline
\end{tabular}%
}
\label{ablation_components}
\end{table}

\paragraph{Input Length:} 
We perform an ablation study to see the influence of the input length $n$ for KCN. We test on $n = 1, 3, 5, 10, 40$. From Tab.~\ref{ablation_hyper} We find that if we only use the current frame, i.e. $n = 1$, there will be a limited improvement compared to CaRTS. Once previous frames are provided $n \geq 3$, the improvement becomes obvious. This indicates that the improvement comes from temporal constraints. However, with a further increase in the input length, the performance does not increase. We suppose this is because increasing the input length will also increase the difficulty of the optimization.

\begin{table}[t]
\centering
\caption{Results of the ablation study for input length}
\resizebox{\columnwidth}{!}{%
\begin{tabular}{|c|c|c|c|c|c|c|}
\hline
 $n$ &  1 & 3  & 5 & 10 & 20 & 40\\
\hline
Dice & $91.3 \pm 3.0$ & $93.0 \pm 2.4$ & $92.6 \pm 2.6$ & $ 92.3 \pm 3.2$ &  $91.7 \pm 3.4$ & $92.3 \pm 3.0$ \\
% \hline
% \hline
%  $\lambda_1$ &  0 & 1  & 10 & 100 & 1000 & 10000\\
% \hline
% Dice & $91.1 \pm 2.9$ & $91.3 \pm 3.0$ & $92.6 \pm 2.6$ & $92.8 \pm 2.7$ &  $91.2 \pm 3.3$ & $91.0 \pm 3.0$ \\
% \hline
% \hline
% $\lambda_2$ & 0 & 0.1  & 1 & 10 & 100 & 1000 \\
% \hline
% Dice &  $92.1 \pm 3.0$ & $92.6 \pm 2.6$ & $92.6 \pm 2.6$ & $92.3 \pm 3.0$ &  $91.1 \pm 3.3$ & $88.4 \pm 3.5$ \\
\hline
\end{tabular}
}
\label{ablation_hyper}
\end{table}

\paragraph{Regularization Strength:} We perform an ablation study to see the influence of the regularization strength $\lambda_1,\lambda_2$ in the spatial-temporal regularization. We separately test on $\lambda_1 = 0, 1, 10, 100, 1000, 10000$ when $\lambda_2 = 1$ and $\lambda_2 = 0, 0.1, 1, 10, 100, 1000$ when $\lambda_1 = 10$. From Tab.~\ref{ablation_lambda} we find that when we set either $\lambda_1$ or $\lambda_2$ to $0$, there are performance drops ($1.5$ / $0.5$ dice score) compared to the default setting. When regularization is too strong ($\lambda_1 \geq 1000$, $\lambda_2 \geq 100$), optimization might also become harder.

 \begin{table}[t]
\centering
\caption{Results of the ablation study for regularization strength}
\resizebox{\columnwidth}{!}{%
\begin{tabular}{|c|c|c|c|c|c|c|}
\hline
 $\lambda_1$ &  0 & 1  & 10 & 100 & 1000 & 10000\\
\hline
Dice & $91.1 \pm 2.9$ & $91.3 \pm 3.0$ & $92.6 \pm 2.6$ & $92.8 \pm 2.7$ &  $91.2 \pm 3.3$ & $91.0 \pm 3.0$ \\
\hline
\hline
$\lambda_2$ & 0 & 0.1  & 1 & 10 & 100 & 1000 \\
\hline
Dice &  $92.1 \pm 3.0$ & $92.6 \pm 2.6$ & $92.6 \pm 2.6$ & $92.3 \pm 3.0$ &  $91.1 \pm 3.3$ & $88.4 \pm 3.5$ \\
\hline
% \hline
\end{tabular}
}
\label{ablation_lambda}
\end{table}

% \paragraph{Overfiting on time-invariant factor:}

% We perform an ablation study to show an insight into the overfitting for CaRTS of a time-invariant factor -- base configuration. We replace kinematics jointly with base configuration and show results with different iteration numbers $k = 0, 1, 3, 5, 10, 30$. Tab.~\ref{ablation_time_inv} shows that when $k$ is small optimization increases the performance. However, as the iteration number per frame increases to $30$, the performance deteriorates. This indicates that it overfits some frames and does not generalize to the whole sequence.

% \begin{table}[t]
% \centering
% \caption{Result for optimizing base configuration using CaRTS}
% \resizebox{\columnwidth}{!}{%
% \begin{tabular}{|c|c|c|c|c|c|c|}
% \hline
%  Iterations & 0 & 1 & 3  & 5 & 10 & 30\\
% \hline
% Dice & $88.4 \pm 2.8$ & $89.7 \pm 2.8$ & $91.3 \pm 2.9$ & $92.0 \pm 2.8$ & $92.5 \pm 2.9$ & $66.9 \pm 21.2$   \\
% \hline
% % \hline
% \end{tabular}
% }
% \label{ablation_time_inv}
% \end{table}

\paragraph{Effectiveness over time:}

\begin{figure}[t]%
\centering
\includegraphics[width=\textwidth]{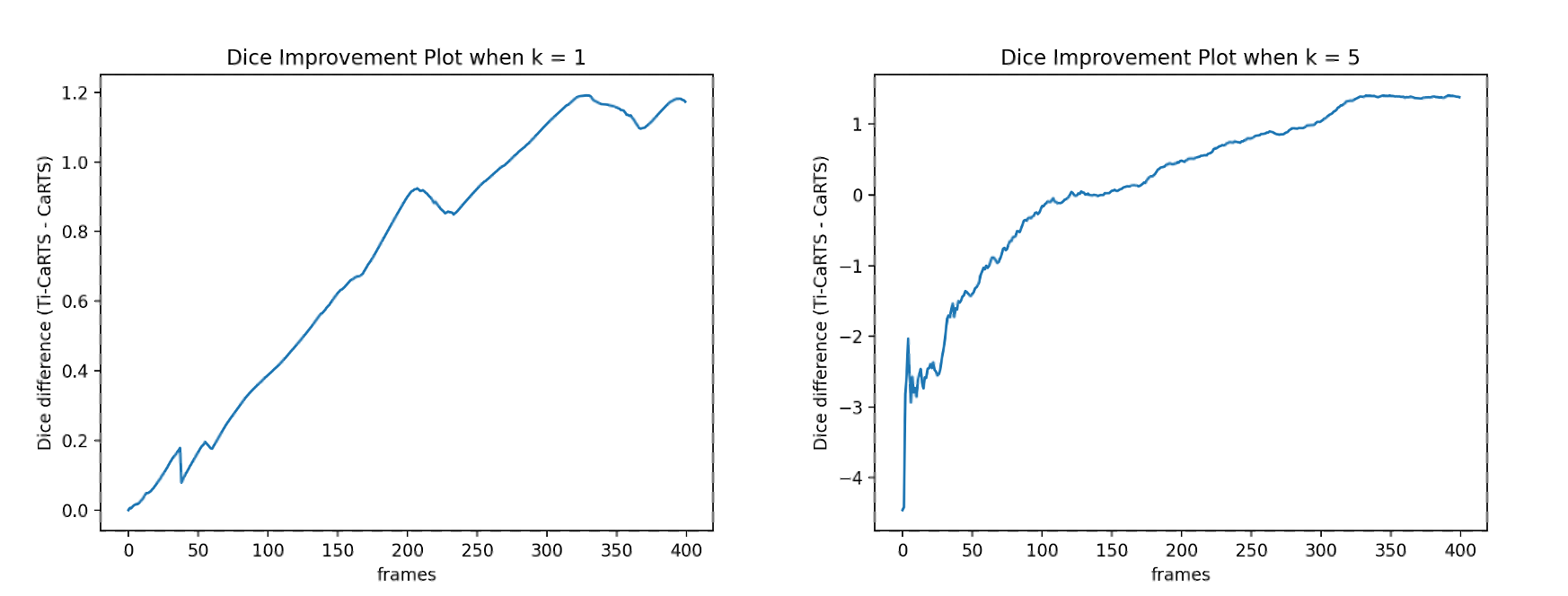}
\caption{Dice improvement over frames}
\label{dice_improvement}
\end{figure}

We perform an ablation study to show an insight into the \ours{}'s effectiveness over time. We calculate the average Dice difference of the first $m$ frames for $m = 1, 2, \dots, 400$ between \ours{} and CaRTS. The result plots are shown in Fig.~\ref{dice_improvement}. As the plots show, the Dice difference increases as $m$ increases. This indicates that with more temporal information having been processed, \ours{}'s advantage with temporal constraints becomes more obvious. We also find that when $m$ is small ($ \leq 50$),  \ours{} performs worse than CaRTS when $k = 5$. This indicates that \ours{} might overfit some early frames. However, when more $m$ increases, \ours{} start to outperform CaRTS.

\section{Limitations}\label{sec5}

Although \ours{} is proven to be effective. There are still limitations that are not fully resolved. Firstly, the challenge to achieve real-time remains. On the one hand, differentiable rendering and backpropagation require more computation than a single feed-forward network. On the other hand, redundant feature extraction operations on similar rendered images also restrict the inference speed. Secondly, the architecture works under the assumption that there is no occlusion and interaction. To deal with occlusion and interaction, information for the environmental factor $\mathbf{E}$ is necessary. The representation of $\mathbf{E}$ might be estimated through vision or other sensors. All of these limitations also imply essential directions for future work.

\section{Conclusion}\label{sec6}

In summary, limited observability causes slow convergence for CaRTS. We propose a temporal causal model and explore underlying temporal constraints in this model. Inspired by the temporal causal model, we propose \ours{} with three novel modules to complement CaRTS-- temporal optimization pipeline, kinematics correction network, and spatial-temporal regularization. \ours{} requires fewer iterations to achieve the same or better performance as CaRTS while achieving the same or better performance in different domains compared to CaRTS. Ablation studies indicate that all modules are effective and the effectiveness comes from temporal constraints.

\paragraph{\textbf{Acknowledgement:}} This research is supported by a collaborative research agreement with the MultiScale Medical Robotics Center at The Chinese University of Hong Kong.

\bibliography{sn-bibliography}% common bib file


\end{document}